# Stochastic Variance-Reduced Cubic Regularized Newton Method

Dongruo Zhou* and Pan Xu† and Quanquan Gu‡

**Abstract**

We propose a stochastic variance-reduced cubic regularized Newton method for non-convex optimization. At the core of our algorithm is a novel semi-stochastic gradient along with a semi-stochastic Hessian, which are specifically designed for cubic regularization method. We show that our algorithm is guaranteed to converge to an $(\epsilon, \sqrt{\epsilon})$-approximately local minimum within $\widetilde{O}(n^{4/5}/\epsilon^{3/2})$ second-order oracle calls, which outperforms the state-of-the-art cubic regularization algorithms including subsampled cubic regularization. Our work also sheds light on the application of variance reduction technique to high-order non-convex optimization methods. Thorough experiments on various non-convex optimization problems support our theory.

## 1 Introduction

We study the following finite-sum optimization problem:

$$\min_{\mathbf{x}\in\mathbb{R}^d} F(\mathbf{x}) = \frac{1}{n}\sum_{i=1}^{n} f_i(\mathbf{x}), \tag{1.1}$$

where $F(\mathbf{x})$ and each $f_i(\mathbf{x})$ can be non-convex. Such problems are common in machine learning, where each $f_i(\mathbf{x})$ is a loss function on a training example (LeCun et al., 2015). Since $F(\mathbf{x})$ is non-convex, finding its global minimum is generally NP-Hard (Hillar and Lim, 2013). As a result, one possible goal is to find an approximate first-order stationary point ($\epsilon-$stationary point):

$$\|\nabla F(\mathbf{x})\| \leq \epsilon,$$

for some given $\epsilon > 0$. A lot of studies have been devoted to this problem including gradient descent (GD), stochastic gradient descent (SGD) (Robbins and Monro, 1951), and their extensions

*Department of Systems and Information Engineering, University of Virginia, Charlottesville, VA 22904, USA; e-mail: `dz4bd@virginia.edu`

†Department of Computer Science, University of Virginia, Charlottesville, VA 22904, USA; e-mail: `px3ds@virginia.edu`

‡Department of Computer Science, University of Virginia, Charlottesville, VA 22904, USA; e-mail: `qg5w@virginia.edu`

(Ghadimi and Lan, 2013; Reddi et al., 2016a; Allen-Zhu and Hazan, 2016; Ghadimi and Lan, 2016). Nevertheless, first-order stationary points can be non-degenerate saddle points or even local maximum in non-convex optimization, which are undesirable. Therefore, a more reasonable objective is to find an approximate second-order stationary point (Nesterov and Polyak, 2006), which is also known as an $(\epsilon_g, \epsilon_h)$-approximate local minimum of $F(\mathbf{x})$:

$$\|\nabla F(\mathbf{x})\|_2 < \epsilon_g, \quad \lambda_{\min}(\nabla^2 F(\mathbf{x})) \geq -\epsilon_h, \tag{1.2}$$

for some given constant $\epsilon_g, \epsilon_h > 0$. In fact, in some machine learning problems like matrix completion (Ge et al., 2016), one finds that every local minimum is a global minimum, suggesting that finding an approximate local minimum is a better choice than a stationary point, and is good enough in many applications. One of the most popular method to achieve this goal is perhaps cubic regularized Newton method, which was introduced by Nesterov and Polyak (2006), and solves the following kind of subproblems in each iteration:

$$\mathbf{h}(\mathbf{x}) = \mathop{\mathrm{argmin}}_{\mathbf{h}\in\mathbb{R}^d} m(\mathbf{h}, \mathbf{x}) = \langle \nabla F(\mathbf{x}), \mathbf{h}\rangle + \frac{1}{2}\langle \nabla^2 F(\mathbf{x})\mathbf{h}, \mathbf{h}\rangle + \frac{\theta}{6}\|\mathbf{h}\|_2^3, \tag{1.3}$$

where $\theta > 0$ is a regularization parameter. Nesterov and Polyak (2006) proved that fixing a starting point $\mathbf{x}_0$, and performing the updating rule $\mathbf{x}_t = \mathbf{x}_{t-1} + \mathbf{h}(\mathbf{x}_{t-1})$, the algorithm can output a sequence $\mathbf{x}_i$ that converges to a local minimum provided that the function is Hessian Lipschitz. However, it can be seen that to solve the subproblem (1.3), one needs to calculate the full gradient $\nabla F(\mathbf{x})$ and Hessian $\nabla^2 F(\mathbf{x})$, which is a big overhead in large scale machine learning problem because $n$ is often very large.

Some recent studies presented various algorithms to avoid the calculation of full gradient and Hessian in cubic regularization. Kohler and Lucchi (2017) used subsampling technique to get approximate gradient and Hessian instead of exact ones, and Xu et al. (2017b) also used subsampled Hessian. Both of them can reduce the computational complexity in some circumstance. However, just like other sampling-based algorithm such as subsampled Newton method Erdogdu and Montanari (2015); Xu et al. (2016); Roostakhorasani and Mahoney (2016a,b); Ye et al. (2017), their convergence rates are worse than that of the Newton method, especially when one needs a high-accuracy solution (i.e., the optimization error $\epsilon$ is small). This is because the subsampling size one needs to achieve certain accuracy may be even larger than the full sample size $n$. Therefore, a natural question arises as follows:

*When we need a high-accuracy local minimum, is there an algorithm that can output an approximate local minimum with better second-order oracle complexity than cubic regularized Newton method?*

In this paper, we give an affirmative answer to the above question. We propose a novel cubic regularization algorithm named Stochastic Variance-Reduced Cubic regularization (**SVR Cubic**), which incorporates the variance reduction techniques (Johnson and Zhang, 2013; Xiao and Zhang, 2014; Allen-Zhu and Hazan, 2016; Reddi et al., 2016a) into the cubic-regularized Newton method. The key component in our algorithm is a novel semi-stochastic gradient, together with a semi-stochastic Hessian, that are specifically designed for cubic regularization. Furthermore, we prove that, for Hessian Lipschitz functions, to attain an approximate $(\epsilon, \sqrt{\rho\epsilon})$-local minimum, our proposed algorithm requires $O(n+n^{4/5}/\epsilon^{3/2})$ Second-order Oracle (SO) calls and $O(1/\epsilon^{3/2})$ Cubic Subproblem Oracle (CSO) calls. Here an ISO oracle represents an evaluation of triple $(f_i(\mathbf{x}), \nabla f_i(\mathbf{x}), \nabla^2 f_i(\mathbf{x}))$, and a CSO oracle denotes an evaluation of the exact solution (or inexact solution) of the cubic subproblem (1.3). Compared with the original cubic regularization algorithm (Nesterov and Polyak, 2006), which requires $O(n/\epsilon^{3/2})$ ISO calls and $O(1/\epsilon^{3/2})$ CSO calls, our proposed algorithm reduces the SO calls by a factor of $\Omega(n^{1/5})$. We also carry out experiments on real data to demonstrate the superior performance of our algorithm.

Our major contributions are summarized as follows:

- We present a novel cubic regularization method with improved oracle complexity. To the best of our knowledge, this is the first algorithm that outperforms cubic regularization without any loss in convergence rate. This is in sharp contrast to subsampled cubic regularization methods (Kohler and Lucchi, 2017; Xu et al., 2017a), which suffer from worse convergence rates than cubic regularization.

- We also extend our algorithm to the case with inexact solution to the cubic regularization subproblem. Similar to previous work (Cartis et al., 2011; Xu et al., 2017a), we also layout a set of sufficient conditions, under which the output of the inexact algorithm is still guaranteed to have the same convergence rate and oracle complexity as the exact algorithm. This further sheds light on the practical implementation of our algorithm.

- As far as we know, our work is the first work, which rigorously demonstrates the advantage of variance reduction for second-order optimization algorithms. Although there exist a few studies (Lucchi et al., 2015; Moritz et al., 2016; Rodomanov and Kropotov, 2016) using variance reduction to accelerate Newton method, none of them can deliver faster rates of convergence than standard Newton method.

**Notation** We use $[n]$ to denote the index set $\{1, 2, \ldots, n\}$. We use $\|\mathbf{v}\|_2$ to denote vector Euclidean norm. For symmetric matrix $\mathbf{H} \in \mathbb{R}^{d\times d}$, we denote its eigenvalues by $\lambda_1(\mathbf{H}) \leq \ldots \leq \lambda_d(\mathbf{H})$, its spectral norm by $\|\mathbf{H}\|_2 = \max\{|\lambda_1(\mathbf{H})|, |\lambda_d(\mathbf{H})|\}$, and the Schatten $r$-norm by $\|\mathbf{H}\|_{S_r} = (\sum_{i=1}^d |\lambda_i(\mathbf{H})|^r)^{1/r}$ for $r \geq 1$. We denote $\mathbf{A} \succeq \mathbf{B}$ if $\lambda_1(\mathbf{A}-\mathbf{B}) \geq 0$ for symmetric matrices $\mathbf{A}, \mathbf{B} \in \mathbb{R}^{d\times d}$. We also note that $\|\mathbf{A}-\mathbf{B}\|_2 \leq C \Rightarrow \|\mathbf{A}\|_2 \succeq \|\mathbf{B}\|_2 - C \cdot \mathbf{I}, C > 0$. We call $\xi$ a Rademacher random variable if $\mathbb{P}(\xi = 1) = \mathbb{P}(\xi = -1) = 1/2$. We use $f_n = O(g_n)$ to denote that $f_n \leq C g_n$ for some constant $C > 0$ and use $f_n = \widetilde{O}(g_n)$ to hide the logarithmic terms of $g_n$.

## 2 Related Work

In this section, we briefly review the relevant work in the literature.

The most related work to ours is the cubic regularized Newton method, which was originally proposed in Nesterov and Polyak (2006). Cartis et al. (2011) presented an adaptive framework of cubic regularization, which uses an adaptive estimation of the local Lipschitz constant and approximate solution to the cubic subproblem. To overcome the computational burden of gradient and Hessian matrix evaluations, Kohler and Lucchi (2017); Xu et al. (2017b,a) proposed to use subsampled gradient and Hessian in cubic regularization. On the other hand, in order to solve the cubic subproblem (1.3) more efficiently, Carmon and Duchi (2016) proposed to use gradient descent, while Agarwal et al. (2017) proposed a sophisticated algorithm based on approximate matrix inverse and approximate PCA. Tripuraneni et al. (2017) proposed a refined stochastic cubic regularization algorithm based on above subproblem solver. However, none of the aforementioned variants of cubic regularization outperforms the original cubic regularization method in terms of the oracle complexity.

Another important line of related research is the variance reduction method, which has been extensively studied for large-scale finite-sum optimization problems. Variance reduction was first proposed in convex finite-sum optimization (Roux et al., 2012; Johnson and Zhang, 2013; Xiao and Zhang, 2014; Defazio et al., 2014), which uses semi-stochastic gradient to reduce the variance of the stochastic gradient and improves the gradient complexity of both stochastic gradient descent (SGD) and gradient descent (GD). Representative algorithms include Stochastic Average Gradient (SAG) (Roux et al., 2012), Stochastic Variance Reduced Gradient (SVRG) (Johnson and Zhang, 2013) and SAGA (Defazio et al., 2014), to mention a few. For non-convex finite-sum optimization, Garber and Hazan (2015); Shalev-Shwartz (2016) proposed algorithms for the setting where each individual function may be non-convex, but their sum is still convex. Later on, Reddi et al. (2016a) and Allen-Zhu and Hazan (2016) extended the SVRG algorithm to the general non-convex finite-sum optimization, which outperforms SGD and GD in terms of gradient complexity as well. However, to the best of our knowledge, it is still an open problem whether variance reduction can also improve the oracle complexity of second-order optimization algorithms.

Last but not the least is the line of research which aims to escape from nondegenerated saddle points by finding the negative curvature direction. There is a vast literature which focuses on algorithms escaping from saddle point by using information of gradient and negative curvature instead of considering the subproblem (1.3). Ge et al. (2015), Jin et al. (2017a) showed that simple (stochastic) gradient descent with perturbation can escape from saddle points. Carmon et al. (2016); Royer and Wright (2017); Allen-Zhu (2017) showed that by calculating the negative curvature using Hessian information, one can find $(\epsilon, \sqrt{\epsilon})$-local minimum faster than the first-order methods. Recent work (Allen-Zhu and Li, 2017; Jin et al., 2017b; Xu et al., 2017c) proposed first-order algorithms that can escape from saddle points without using Hessian information.

For better comparison of our algorithm with the most related algorithms in terms of SO and CSO oracle complexities, we summarize the results in Table 1. It can be seen from Table 1 that our algorithm (SVR Cubic) achieves the lowest (SO and CSO) oracle complexity compared with the original cubic regularization method (Nesterov and Polyak, 2006) which employs full gradient and Hessian evaluations and the subsampled cubic method (Kohler and Lucchi, 2017; Xu et al., 2017b). In particular, our algorithm reduces the SO oracle complexity of cubic regularization by a factor of $n^{1/5}$ for finding an $(\epsilon, \sqrt{\epsilon})$-local minimum. We will provide more detailed discussion in the main theory section.

[1] It is the refined rate proved by Xu et al. (2017b) for the subsampled cubic regularization algorithm proposed in Kohler and Lucchi (2017)

Table 1: Comparisons between different methods to find $(\epsilon, \sqrt{\epsilon})$-local minimum on the second-order oracle (SO) complexity and and the cubic subproblem oracle (CSO) complexity.

| Algorithm | SO calls | CSO calls | Gradient Lipschitz | Hessian Lipschitz |
|---|---|---|---|---|
| Cubic regularization (Nesterov and Polyak, 2006) | $O(n/\epsilon^{3/2})$ | $O(1/\epsilon^{3/2})$ | no | yes |
| Subsampled cubic regularization (Kohler and Lucchi, 2017; Xu et al., 2017b) | $\widetilde{O}(n/\epsilon^{3/2} + 1/\epsilon^{5/2})^1$ | $O(1/\epsilon^{3/2})$ | yes | yes |
| SVR Cubic (this paper) | $\widetilde{O}(n + n^{4/5}/\epsilon^{3/2})$ | $O(1/\epsilon^{3/2})$ | no | yes |

# 3 The Proposed Algorithm

In this section, we present a novel algorithm, which utilizes stochastic variance reduction techniques to improve cubic regularization method.

To reduce the computation burden of gradient and Hessian matrix evaluations in the cubic regularization updates in (1.3), subsampled gradient and Hessian matrix have been used in subsampled cubic regularization (Kohler and Lucchi, 2017; Xu et al., 2017b) and stochastic cubic regularization (Tripuraneni et al., 2017). Nevertheless the stochastic gradient and Hessian matrix have large variances, which undermine the convergence performance. Inspired by SVRG (Johnson and Zhang, 2013), we propose to use a semi-stochastic version of gradient and Hessian matrix, which can control the variances automatically. Specifically, our algorithm has two loops. At the beginning of the $s$-th iteration of the outer loop, we denote $\widehat{\mathbf{x}}^s = \mathbf{x}_0^{s+1}$. We first calculate the full gradient $\mathbf{g}^s = \nabla F(\widehat{\mathbf{x}}^s)$ and Hessian matrix $\mathbf{H}^s = \nabla^2 F(\widehat{\mathbf{x}}^s)$, which are stored for further references in the inner loop. At the $t$-th iteration of the inner loop, we calculate the following semi-stochastic gradient and Hessian matrix:

$$\mathbf{v}_t^{s+1} = \frac{1}{b_g} \sum_{i_t \in I_g} \big(\nabla f_{i_t}(\mathbf{x}_t^{s+1}) - \nabla f_{i_t}(\widehat{\mathbf{x}}^s) + \mathbf{g}^s\big) - \frac{1}{b_g} \sum_{i_t \in I_g} \big(\nabla^2 f_{i_t}(\widehat{\mathbf{x}}^s) - \mathbf{H}^s\big)\big(\mathbf{x}_t^{s+1} - \widehat{\mathbf{x}}^s\big), \tag{3.1}$$

$$\mathbf{U}_t^{s+1} = \frac{1}{b_h} \sum_{j_t \in I_h} \big(\nabla^2 f_{j_t}(\mathbf{x}_t^{s+1}) - \nabla^2 f_{j_t}(\widehat{\mathbf{x}}^s)\big) + \mathbf{H}^s, \tag{3.2}$$

where $I_g$ and $I_h$ are batch index sets, and the batch sizes will be decided later. In each inner iteration, we solve the following cubic regularization subproblem:

$$\mathbf{h}_t^{s+1} = \operatorname{argmin} m_t^{s+1}(\mathbf{h}) = \langle \mathbf{v}_t^{s+1} \mathbf{h} \rangle + \frac{1}{2} \langle \mathbf{U}_t^{s+1} \mathbf{h}, \mathbf{h} \rangle + \frac{M_{s+1,t}}{6} \|\mathbf{h}\|_2^3. \tag{3.3}$$

Then we perform the update $\mathbf{x}_{t+1}^{s+1} = \mathbf{x}_t^{s+1} + \mathbf{h}_t^{s+1}$ in the $t$-th iteration of the inner loop. The proposed algorithm is displayed in Algorithm 1.

**Algorithm 1** Stochastic Variance Reduction Cubic Regularization (**SVR Cubic**)

**Input:** batch size $b_g, b_h$, penalty parameter $M_{s,t}$, $s = 1 \dots S, t = 0 \dots T$, starting point $\widehat{\mathbf{x}}^1$.
**Initialization**
**for** $s = 1, \dots, S$ **do**
  $\mathbf{x}_0^{s+1} = \widehat{\mathbf{x}}^s$
  $\mathbf{g}^s = \nabla F(\widehat{\mathbf{x}}^s) = \frac{1}{n}\sum_{i=1}^n \nabla f_i(\widehat{\mathbf{x}}^s), \mathbf{H}^s = \frac{1}{n}\sum_{i=1}^n \nabla^2 f_i(\widehat{\mathbf{x}}^s)$
  **for** $t = 0, \dots, T-1$ **do**
    Sample index set $I_g, I_h, |I_g| = b_g, |I_h| = b_h$;
    $\mathbf{v}_t^{s+1} = \frac{1}{b_g}\sum_{i_t \in I_g} \nabla f_{i_t}(\mathbf{x}_t^{s+1}) - \nabla f_{i_t}(\widehat{\mathbf{x}}^s) + \mathbf{g}^s - \big(\frac{1}{b_g}\sum_{i_t \in I_g} \nabla^2 f_{i_t}(\widehat{\mathbf{x}}^s) - \mathbf{H}^s\big)(\mathbf{x}_t^{s+1} - \widehat{\mathbf{x}}^s)$
    $\mathbf{U}_t^{s+1} = \frac{1}{b_h}\big(\sum_{j_t \in I_h} \nabla^2 f_{j_t}(\mathbf{x}_t^{s+1}) - \nabla^2 f_{j_t}(\widehat{\mathbf{x}}^s)\big) + \mathbf{H}^s$
    $\mathbf{h}_t^{s+1} = \mathrm{argmin}_{\mathbf{h}} \langle \mathbf{v}_t^{s+1}, \mathbf{h} \rangle + \frac{1}{2}\langle \mathbf{U}_t^{s+1}\mathbf{h}, \mathbf{h}\rangle + \frac{M_{s+1,t}}{6}\|\mathbf{h}\|_2^3,$
    $\mathbf{x}_{t+1}^{s+1} = \mathbf{x}_t^{s+1} + \mathbf{h}_t^{s+1}$
  **end for**
  $\widehat{\mathbf{x}}^{s+1} = \mathbf{x}_T^{s+1}$
**end for**
**Output:** random choose one $\mathbf{x}_t^s$, for $t = 0, ..., T$ and $s = 1, ..., S$.

There are two notable features of our "estimator" of the full gradient and Hessian in each inner loop, compared with that used in SVRG (Johnson and Zhang, 2013). The first is that our gradient and Hessian estimators consist of mini-batches of stochastic gradient and Hessian. The second one is that we use second order information when we construct the gradient estimator $\mathbf{v}_t^{s+1}$, while classical SVRG only uses first order information to build it. Intuitively speaking, both features are used to make a more accurate estimation of the true gradient and Hessian with affordable oracle calls. Note that similar approximations of the gradient and Hessian matrix have been staged in recent work by Gower et al. (2017) and Wai et al. (2017), where they used this new kind of estimator for traditional SVRG in the convex setting, which radically differs from our setting.

# 4 Main Theory

We first lay down the following Hessian Lipschitz assumption, which are necessary for our analysis and are widely used in the literature (Nesterov and Polyak, 2006; Xu et al., 2016; Kohler and Lucchi, 2017).

**Assumption 4.1** (Hessian Lipschitz)**.** There exists a constant $\rho > 0$, such that for all $\mathbf{x}, \mathbf{y}$ and $i \in [n]$

$$\|\nabla^2 f_i(\mathbf{x}) - \nabla^2 f_i(\mathbf{y})\|_2 \le \rho \|\mathbf{x} - \mathbf{y}\|_2.$$

In fact, this is the only assumption we need to prove our theoretical results. The Hessian Lipschitz assumption plays a central role in controlling the changing speed of second order information. It is obvious that Assumption 4.1 implies the Hessian Lipschitz assumption of $F$, which, according to Nesterov and Polyak (2006), is also equivalent to the following lemma.

**Lemma 4.2.** Let function $F : \mathbf{x} \to \mathbb{R}^d$ satisfy $\rho$-Hessian Lipschitz assumption, then for any $\mathbf{h} \in \mathbb{R}^d$, it holds that

$$\|\nabla^2 F(\mathbf{x}) - \nabla^2 F(\mathbf{y})\|_2 \le \rho\|\mathbf{x} - \mathbf{y}\|_2,$$
$$F(\mathbf{x} + \mathbf{h}) \le F(\mathbf{x}) + \langle \nabla F(\mathbf{x}), \mathbf{h}\rangle + \frac{1}{2}\langle \nabla^2 F(\mathbf{x})\mathbf{h}, \mathbf{h}\rangle + \frac{\rho}{6}\|\mathbf{h}\|_2^3,$$
$$\|\nabla F(\mathbf{x} + \mathbf{h}) - \nabla F(\mathbf{x}) - \nabla^2 F(\mathbf{x})\mathbf{h}\|_2 \le \frac{\rho}{2}\|\mathbf{h}\|_2^2.$$

We then define the following optimal function gap between initial point $\mathbf{x}_0$ and the global minimum of $F$.

**Definition 4.3** (Optimal Gap)**.** For function $F(\cdot)$ and the initial point $\mathbf{x}_0$, let $\Delta_F$ be

$$\Delta_F = \inf\{\Delta \in \mathbb{R} : F(\mathbf{x}_0) - F^* \le \Delta\},$$

where $F^* = \inf_{\mathbf{x}\in\mathbb{R}^d} F(\mathbf{x})$.

Without loss of generality, we assume $\Delta_F < +\infty$ throughout this paper.

Before we present nonasympotic convergence results of Algorithm 1, we define

$$\mu(\mathbf{x}_t^{s+1}) = \max\left\{\|\nabla F(\mathbf{x}_t^{s+1})\|_2^{3/2}, -\frac{\lambda_{\min}^3(\nabla^2 F(\mathbf{x}_t^{s+1}))}{[M_{s+1,t}]^{3/2}}\right\}. \tag{4.1}$$

By definition in (4.1), $\mu(\mathbf{x}_t^{s+1}) < \epsilon^{3/2}$ holds if and only if

$$\|\nabla F(\mathbf{x}_t^{s+1})\|_2 \le \epsilon, \quad \lambda_{\min}\big(\nabla^2 F(\mathbf{x}_t^{s+1})\big) > -\sqrt{M_{s+1,t}\epsilon}. \tag{4.2}$$

Therefore, in order to find an $(\epsilon, \sqrt{\rho\epsilon})$-local minimum of the non-convex function $F$, it suffices to find a point $\mathbf{x}_t^{s+1}$ which satisfies $\mu(\mathbf{x}_t^{s+1}) < \epsilon^{3/2}$, and $M_{s+1,t} = O(\rho)$ for all $s, t$. Next we define our oracles formally:

**Definition 4.4** (Second-order Oracle)**.** Given an index $i$ and a point $\mathbf{x}$, one second-order oracle (SO) call returns such a triple:

$$[f_i(\mathbf{x}), \nabla f_i(\mathbf{x}), \nabla^2 f_i(\mathbf{x})]. \tag{4.3}$$

**Definition 4.5** (Cubic Subproblem Oracle)**.** Given a vector $\mathbf{g} \in \mathbb{R}^d$, a Hessian matrix $\mathbf{H}$ and a positive constant $\theta$, one Cubic Subproblem Oracle (CSO) call returns $\mathbf{h}_{\text{sol}}$, where $\mathbf{h}_{\text{sol}}$ can be solved exactly as follows

$$\mathbf{h}_{\text{sol}} = \underset{\mathbf{h}\in\mathbb{R}^d}{\text{argmin}} \langle \mathbf{g}, \mathbf{h}\rangle + \frac{1}{2}\langle \mathbf{h}, \mathbf{H}\mathbf{h}\rangle + \frac{\theta}{6}\|\mathbf{h}\|_2^3.$$

**Remark 4.6.** The second-order oracle is a special form of *Information Oracle* which is introduced by Nesterov, which returns gradient, Hessian and all high order derivatives of objective function $F(\mathbf{x})$. Here, our second-order oracle will only returns first and second order information at some point of single objective $f_i$ instead of $F$. We argue that it is a reasonable adaption because in this paper we focus on finite-sum objective function. The Cubic Subproblem Oracle will return an exact or inexact solution of (3.3), which plays an important role in both theory and practice.

Now we are ready to give a general convergence result of Algorithm 1:

**Theorem 4.7.** Let $A_t, B_t, \alpha_t$ and $\beta_t$ be arbitrary positive constants, choose $M_{s,t} = M_t$ for each $s$. Define parameter sequences $\{\Theta_t\}_{t=0}^T$ and $\{c_t\}_{t=0}^T$ as follows

$$\begin{aligned} c_t &= \left(\frac{\Theta_t}{M_t^{3/2}} + \frac{1}{A_t^{1/2}}\right)\frac{\rho^{3/2}}{b_g^{3/4}} + \left(\frac{\Theta_t}{M_t^3} + \frac{1}{B_t^2}\right) \cdot \frac{C\rho^3(\log d)^{3/2}}{b_h^{3/2}} + c_{t+1}\left(1 + \frac{1}{\alpha_t^2} + \frac{2}{\beta_t^{1/2}}\right), \\ \Theta_t &= \frac{3M_t - 2\rho - 4A_t - 4B_t}{12} - c_{t+1}(1 + 2\alpha_t + \beta_t), \\ c_T &= 0, \end{aligned} \tag{4.4}$$

where $\rho$ is the Hessian Lipschitz constant, $M_t$ is the regularization parameter of Algorithm 1, and $C$ is an absolute constant. If setting batch size $b_h > 25 \log d$, $M_t = O(\rho)$, and $\Theta_t > 0$ for all $t$, then the output of Algorithm 1 satisfies

$$\mathbb{E}[\mu(\mathbf{x}_{\text{out}})] \le \frac{\mathbb{E}\big[F(\widehat{\mathbf{x}}^0) - F^*\big]}{\gamma_n ST}, \tag{4.5}$$

where $\gamma_n = \min_t \Theta_t/(15M_t^{3/2})$.

**Remark 4.8.** To ensure that $\mathbf{x}_{\text{out}}$ is an $(\epsilon, \sqrt{\rho\epsilon})$-local minimum, we can set the right hand side of (4.5) to be less then $\epsilon^{3/2}$. This immediately implies that the total iteration complexity of Algorithm 1 is $ST = O(\mathbb{E}\big[F(\widehat{\mathbf{x}}^0) - F^*\big]/\epsilon^{3/2})$, which matches the iteration complexity of cubic regularization Nesterov and Polyak (2006).

**Remark 4.9.** Note that there is a $\log d$ term in the expression of parameter $c_t$, and it is only related to Hessian batch size $b_h$. The $\log d$ term comes from matrix concentration inequalities, which is believed to be unavoidable (Tropp et al., 2015). In other words, the batch size of Hessian matrix $b_h$ has a inevitable relation to dimension $d$, unlike the batch size of gradient $b_g$.

The iteration complexity result in Theorem 4.7 depends on a series of parameter defined as in (4.4). In the following corollary, we will show how to choose these parameters in practice to achieve a better oracle complexity.

**Corollary 4.10.** Let batch sizes $b_g$ and $b_h$ satisfy $\sqrt{b_g} = b_h/\log d = 1400n^{2/5}$. Set the parameters in Theorem 4.7 as follows

$$A_t = B_t = 125\rho, \alpha_t = \sqrt{2}n^{1/10}, \beta_t = 4n^{2/5}.$$

$\Theta_t$ and $c_t$ are defined as in (4.4). Let the cubic regularization parameter be $M_t = 2000\rho$, and the epoch length be $T = n^{1/5}$. Then Algorithm 1 converges to a $(\epsilon, \sqrt{\rho\epsilon})$-local minimum with

$$O\left(n + \frac{\Delta_F\sqrt{\rho}n^{4/5}}{\epsilon^{3/2}}\right) \text{ SO calls and } O\left(\frac{\Delta_F\sqrt{\rho}}{\epsilon^{3/2}}\right) \text{ CSO calls.} \tag{4.6}$$

**Remark 4.11.** Corollary 4.10 states that we can reduce the SO calls by setting the batch size $b_g, b_h$ related to $n$. In contrast, in order to achieve a $(\epsilon, \sqrt{\rho\epsilon})$ local minimum, original cubic regularization method in Nesterov and Polyak (2006) needs $O(n/\epsilon^{3/2})$ second-order oracle calls, which is by a factor of $n^{1/5}$ worse than ours. And subsampled cubic regularization Kohler and Lucchi (2017); Xu et al. (2017b) requires $\widetilde{O}(n/\epsilon^{3/2} + 1/\epsilon^{5/2})$ SO calls, which is even worse.

## 5 Practical Algorithm with Inexact Oracle

In practice, the exact solution to the cubic subproblem (3.3) cannot be obtained. Instead, one can only get an approximate solution by some inexact solver. Thus we replace the CSO oracle in (4.5) with the following inexact CSO oracle

$$\widetilde{\mathbf{h}}_{\mathrm{sol}} \approx \operatorname*{argmin}_{\mathbf{h}\in\mathbb{R}^d} \langle \mathbf{g}, \mathbf{h}\rangle + \frac{1}{2}\langle \mathbf{h}, \mathbf{H}\mathbf{h}\rangle + \frac{\theta}{6}\|\mathbf{h}\|_2^3.$$

To analyze the performance of Algorithm 1 with inexact cubic subproblem solver, we replace the exact solver in Line 10 of Algorithm 1 with

$$\widetilde{\mathbf{h}}_t^{s+1} \approx \operatorname{argmin} m_t^{s+1}(\mathbf{h}). \tag{5.1}$$

In order to characterize the above inexact solution, we present the following sufficient condition, under which inexact solution can ensure the same oracle complexity as the exact solution:

**Condition 5.1** (Inexact condition)**.** For each $s, t$ and given $\delta > 0$, $\widetilde{\mathbf{h}}_t^{s+1}$ satisfies $\delta$- inexact condition if $\widetilde{\mathbf{h}}_t^{s+1}$ satisfies

$$\begin{aligned} m_t^{s+1}(\widetilde{\mathbf{h}}_t^{s+1}) &\le -\frac{M_{s+1,t}}{12}\|\widetilde{\mathbf{h}}_t^{s+1}\|_2^3 + \delta, \\ \|\nabla m_t^{s+1}(\widetilde{\mathbf{h}}_t^{s+1}\| &\le \delta^{3/2}, \\ \|\mathbf{h}_t^{s+1}\|_2^3 &\le \|\widetilde{\mathbf{h}}_t^{s+1}\|_2^3 + \delta. \end{aligned}$$

**Remark 5.2.** Similar inexact conditions have been studied in the literature of cubic regularization. For instance, Nesterov and Polyak (2006) presented a practical way to solve the cubic subproblem without termination condition. Cartis et al. (2011); Kohler and Lucchi (2017) presented termination criteria for approximate solution to cubic subproblem, which is slightly different from our condition. In general, the termination criteria in Cartis et al. (2011); Kohler and Lucchi (2017) contains a non-linear equation, which is hard to verify and less practical. In contrast, our inexact condition only contains inequality, which is easy to be verified in practice.

Next we give the convergence result with inexact CSO oracle:

**Theorem 5.3.** Let $\widetilde{\mathbf{h}}_t^{s+1}$ to be the output in each inner loop of Algorithm 1 which satisfies Condition 5.1. Let $A_t, B_t, \alpha_t, \beta_t > 0$ be arbitrary constants. Let $M_{s,t} = M_t$ for each $s$, and $\Theta_t$ and $c_t$ are defined in (4.4), where $1 \le t \le T$. If choosing batch size $b_h > 25\log d$ and $M_t = O(\rho)$, and $\Theta_t > 0$ for all $t$, then the output of Algorithm 1 with inexact subproblem solver satisfies:

$$\mathbb{E}[\mu(\mathbf{x}_{\mathrm{out}})] \le \frac{\mathbb{E}[F(\widehat{\mathbf{x}}^0) - F^*]}{\gamma_n ST} + \delta'_t,$$

where

$$\delta'_t = \delta \cdot \bigg(\Theta_t + \frac{\Theta_t}{2M_t^{3/2}} + 1\bigg), \gamma_n = \min_t \Theta_t/(15M_t^{3/2}).$$

**Remark 5.4.** By the definition of $\mu(\mathbf{x})$ in (4.1) and (4.2), in order to attain an $(\epsilon, \sqrt{\epsilon})$ local minimum, we require $\mathbb{E}[\mu(\mathbf{x}_{\text{out}})] \leq \epsilon^{3/2}$ and thus $\delta_t' < \epsilon^{3/2}$, which implies that $\delta$ in Condition 5.1 should satisfy $\delta < \epsilon^{3/2}/(\Theta_t + \Theta_t/(2M_t^{3/2}) + 1)$. Thus the total iteration complexity of Algorithm 1 is $O(\Delta_F/(\gamma_n \epsilon^{3/2}))$.

By the same choice of parameters, Algorithm 1 with inexact oracle can achieve a reduction in SO calls.

**Corollary 5.5.** Under Condition 5.1, and under the same conditions as in Corollary 4.10, the output of Algorithm 1 with the inexact subproblem solver satisfies $\mathbb{E}\mu(\mathbf{x}_{\text{out}}) \leq \epsilon^{3/2} + \delta_f$ within

$$O\bigg(n + \frac{\Delta_F \sqrt{\rho} n^{4/5}}{\epsilon^{3/2}}\bigg) \text{ SO calls and } O\bigg(\frac{\Delta_F\sqrt{\rho}}{\epsilon^{3/2}}\bigg) \text{ CSO calls,}$$

where $\delta_f = O(\rho\delta)$.

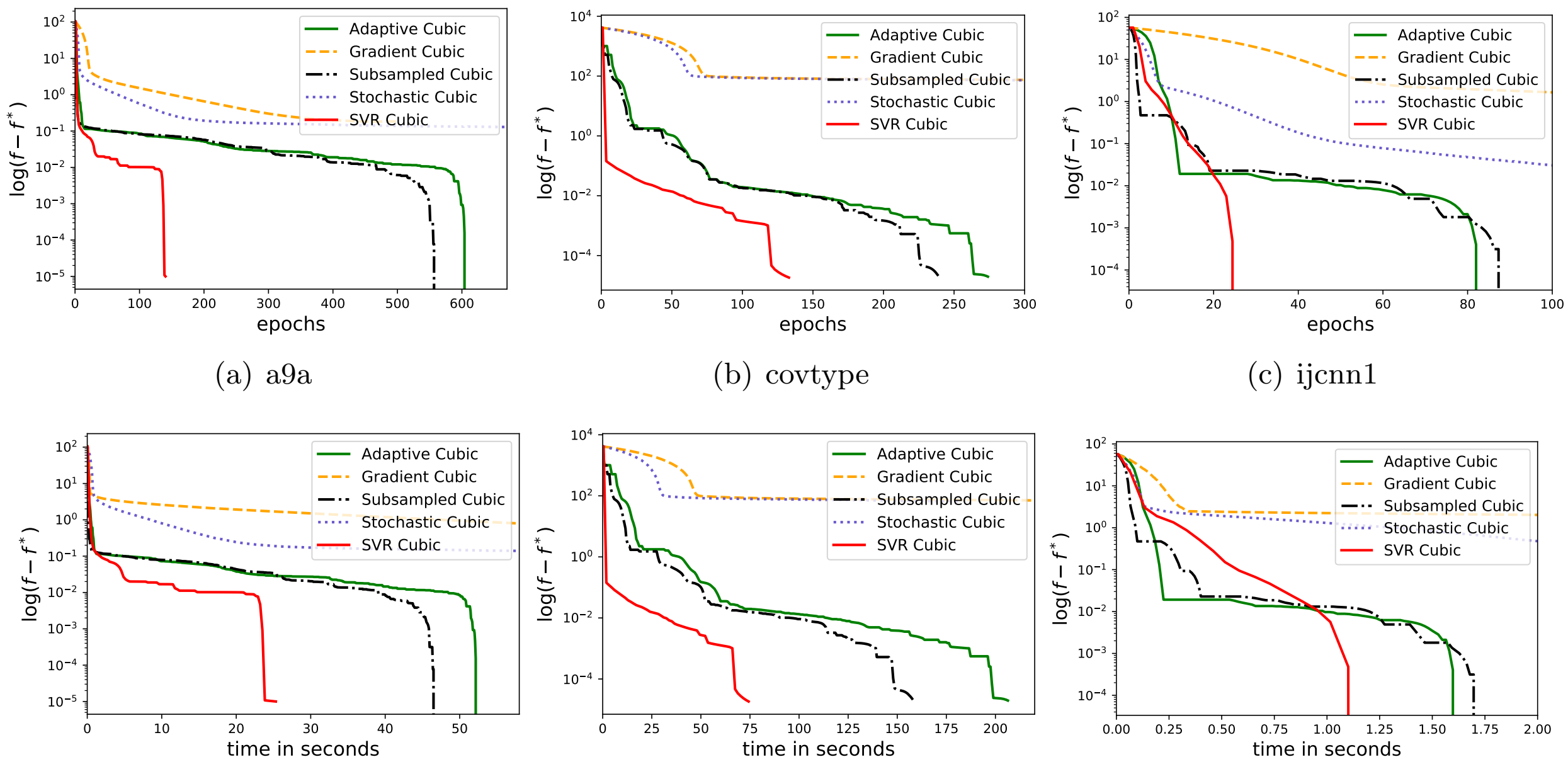

(a) a9a (b) covtype (c) ijcnn1

(d) a9a (e) covtype (f) ijcnn1

Figure 1: Logarithmic function value gap for nonconvex regularized logistic regression on different datasets. (a), (b) and (c) present the oracle complexity comparison; (d), (e) and (f) present the runtime comparison.

# 6 Experiments

In this section, we present numerical experiments on different non-convex Empirical Risk Minimization (ERM) problems and on different datasets to validate the advantage of our proposed algorithm **SVR Cubic** in finding approximate local minima.

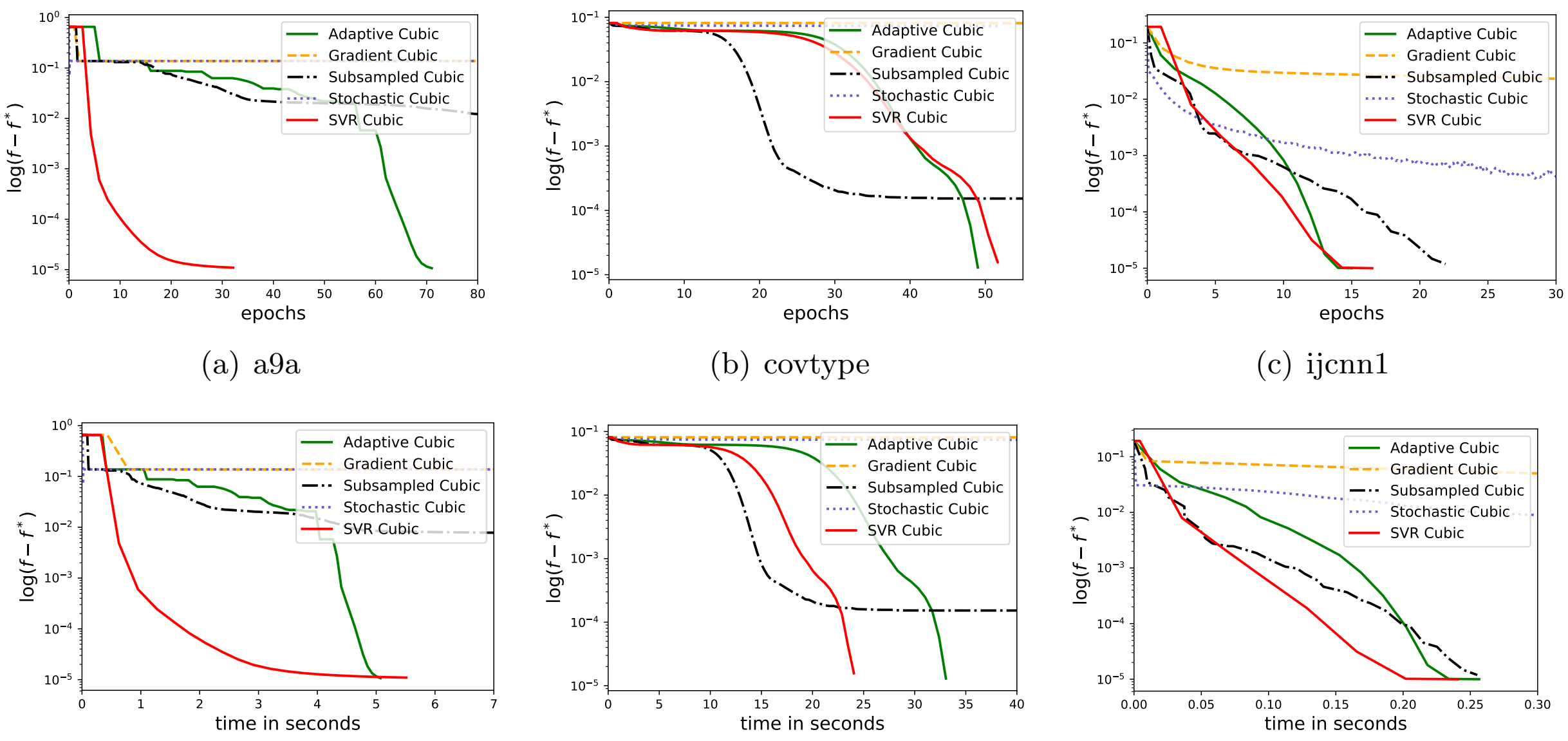

(a) a9a (b) covtype (c) ijcnn1

(d) a9a (e) covtype (f) ijcnn1

Figure 2: Logarithmic function value gap for nonlinear least square on different datasets. (a), (b) and (c) present the oracle complexity comparison; (d), (e) and (f) present the runtime comparison.

**Baselines**: We compare our algorithm with Adaptive Cubic Regularization (**Adaptive Cubic**) (Cartis et al., 2011), Subsampled Cubic Regularization (**Subsampled Cubic**) (Kohler and Lucchi, 2017), Stochastic Cubic Regularization (**Stochastic Cubic**) (Tripuraneni et al., 2017) and Gradient Cubic Regularization (**Gradient Cubic**) (Carmon and Duchi, 2016). All of above algorithms are carefully tuned for a fair comparison.

**The Calculation for SO calls**: Here we list the SO call each algorithm needs for one loop. For **Stochastic Cubic**, each loop costs $(B_g + B_h)$ SO calls, where $B_g$ and $B_h$ to denote the subsampling size of gradient and Hessian. For **Stochastic Cubic**, each loop costs $(n_g + n_h)$ SO calls, where we use $n_g$ and $n_h$ to denote the subsampling size of gradient and Hessian-vector operator. **Gradient Cubic** and **Adaptive Cubic** cost $n$ SO calls in each loop. Finally, we define the amount of epochs is the amount of SO call divided by $n$.

**Parameter tuning and subproblem solver**: For each algorithm and each dataset, we choose different $b_g, b_h, T$ for the best performance. Meanwhile, we also use two different strategies for choosing $M_{s,t}$: the first one is to fix $M_{s,t} = M$ in each iteration, which is proved to enjoy good convergence performance; the other one is to choose $M_{s,t} = \alpha/(1+\beta)^{(s+t/T)}, \alpha, \beta > 0$ for each iteration. This choice of parameter is similar to the choice of penalty parameter in **Subsampled Cubic** and **Adaptive Cubic**, which sometimes makes some algorithms behave better in our experiment. As to the solver for subproblem (3.3) in each loop, we choose to use the Lanczos-type method introduced in Cartis et al. (2011).

**Datasets**: The datasets we use are *a9a, covtype, ijcnn1*, which are common datasets used in ERM problems. The detailed information about these datasets are in Table 2.

**Non-convex regularized logistic regression**: The first nonconvex problem we choose is a binary logistic regression problem with a non-convex regularizer $\sum_{i=1}^{d} \lambda \mathbf{w}_{(i)}^2/(1+\mathbf{w}_{(i)}^2)$ (Reddi et al., 2016b). More specifically, suppose we are given training data $\{\mathbf{x}_i, y_i\}_{i=1}^n$, where $\mathbf{x}_i \in \mathbb{R}^d$ and $y_i \in \{0,1\}$ are feature vectors and labels corresponding to the $i$-th data points. The minimization problem is as follows

$$\min_{\mathbf{w}\in\mathbb{R}^d} \frac{1}{n}\sum_{i=1}^{n} y_i \log \phi(\mathbf{x}_i^T\mathbf{w}) + (1-y_i)\log[1-\phi(\mathbf{x}_i^T\mathbf{w})] + \sum_{i=1}^{d} \lambda \mathbf{w}_{(i)}^2/(1+\mathbf{w}_{(i)}^2),$$

where $\phi(x) = 1/(1+\exp(-x))$ is the sigmoid function. We fix $\lambda = 10$ in our experiments. The experiment results are shown in Figure 1.

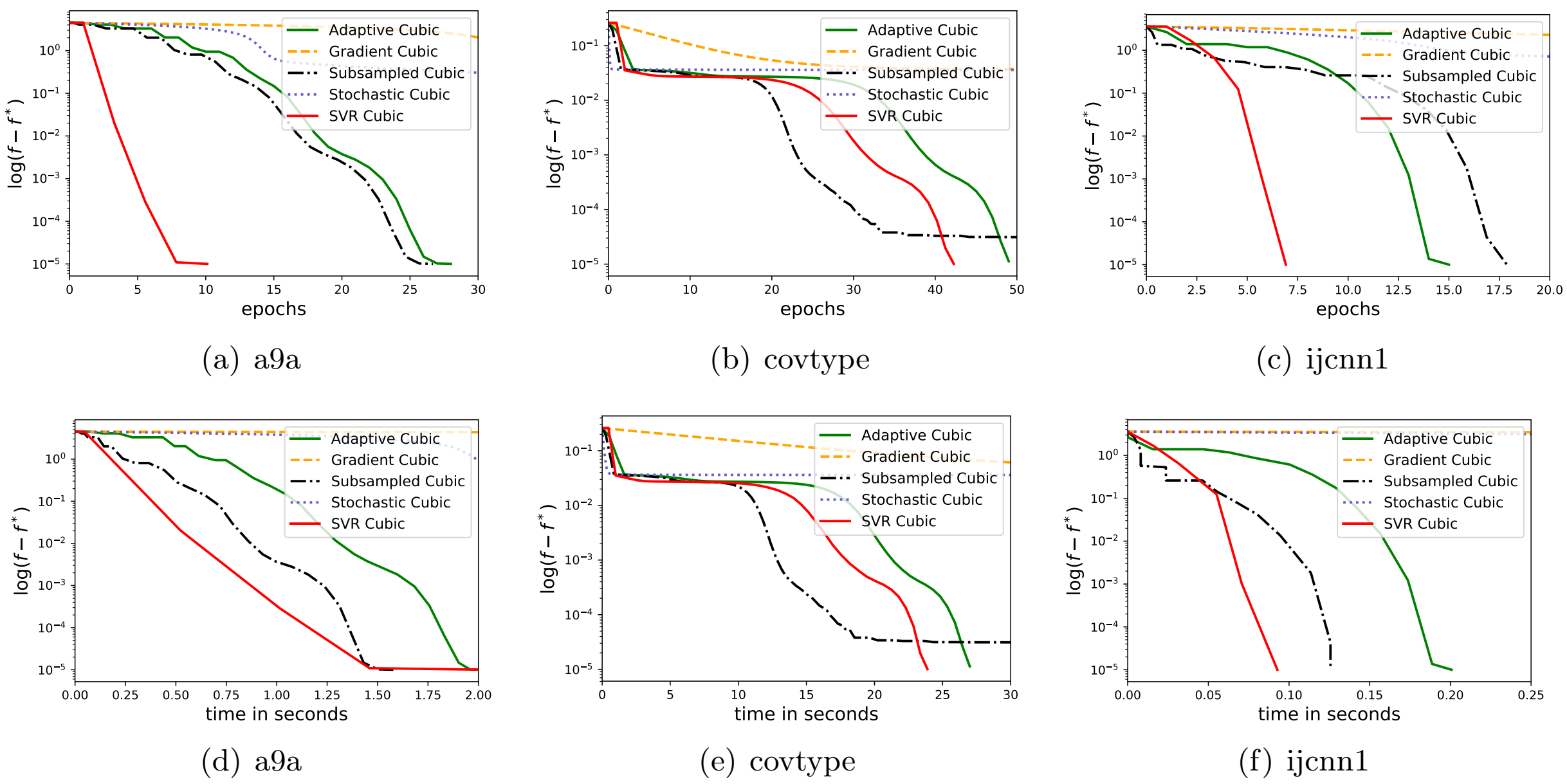

Figure 3: Logarithmic function value gap for robust linear regression on different datasets. (a), (b) and (c) present the oracle complexity comparison; (d), (e) and (f) present the runtime comparison.

Table 2: Overview of the datasets used in our experiments

| Dataset | sample size $n$ | dimension $d$ |
|---|---|---|
| *a9a* | 32,561 | 123 |
| *covtype* | 581,012 | 54 |
| *ijcnn1* | 35,000 | 22 |

**Nonlinear linear squares**: The second problem is a non-linear least squares problem which focuses on the task of binary linear classification (Xu et al., 2017a). Given training data $\{\mathbf{x}_i, y_i\}_{i=1}^n$, where $\mathbf{x}_i \in \mathbb{R}^d$ and $y_i \in \{0,1\}$ are feature vectors and labels corresponding to the $i$-th data points.

The minimization problem is

$$\min_{\mathbf{w}\in\mathbb{R}^d} \frac{1}{n}\sum_{i=1}^{n}[y_i - \phi(\mathbf{x}_i^T\mathbf{w})]^2$$

where $\phi(x) = 1/(1+\exp(-x))$ is the sigmoid function. The experiment results are shown in Figure 2.

**Robust linear regression**: The third problem is a robust linear regression problem where we use a non-convex robust loss function $\log(x^2/2+1)$ (Barron, 2017) instead of square loss in least square regression. Given a training sample $\{\mathbf{x}_i, y_i\}_{i=1}^n$, where $\mathbf{x}_i \in \mathbb{R}^d$ and $y_i \in \{0,1\}$ are feature vectors and labels corresponding to the $i$-th data point. The minimization problem is

$$\min_{\mathbf{w}\in\mathbb{R}^d} \frac{1}{n}\sum_{i=1}^{n}\eta(y_i - \mathbf{x}_i^T\mathbf{w}),$$

where $\eta(x) = \log(x^2/2+1)$. The experimental results are shown in Figure 3.

From Figures 1, 2 and 3, we can see that our algorithm outperforms all the other baseline algorithms on all the datasets. The only exception happens in the non-linear least square problem and the robust linear regression problem on the *covtype* dataset, where our algorithm behaves a little worse than **Adaptive Cubic** at the high accuracy regime in terms of epoch counts. However, under this setting, our algorithm still outperforms the other baselines in terms of the cpu time.

# 7 Conclusions

In this paper, we propose a novel second-order algorithm for non-convex optimization called **SVR Cubic**. Our algorithm is the first algorithm which improves the oracle complexity of cubic regularization and its subsampled variants under certain regime using variance reduction techniques. We also show that similar oracle complexity also holds with inexact oracle. Under both settings our algorithm outperforms the state-of-the-art.